\documentclass{article}

\usepackage{microtype}
\usepackage{graphicx}
\usepackage{subcaption}
\usepackage{booktabs} 

\usepackage{hyperref}

\usepackage{xcolor}
\definecolor{mygreen}{rgb}{0.0, 0.6, 0.0}

\usepackage[preprint]{icml2026}

\usepackage{amsmath}
\usepackage{amssymb}
\usepackage{mathtools}
\usepackage{amsthm}
\usepackage{multirow}
\usepackage{enumitem}
\usepackage{graphicx}
\usepackage{array}
\usepackage{ragged2e}

\usepackage[capitalize,noabbrev]{cleveref}

\theoremstyle{plain}

\theoremstyle{definition}

\theoremstyle{remark}

\usepackage[textsize=tiny]{todonotes}
\usepackage[utf8]{inputenc}
\usepackage[most]{tcolorbox}

\usepackage{titlesec}
\titlespacing*{\paragraph}{0pt}{0.15ex}{1em}
\titlespacing*{\section}{0pt}{1ex}{0.3ex}
\titlespacing*{\subsection}{0pt}{0.5ex}{0.15ex}
\AtBeginDocument{
    \setlength{\abovedisplayskip}{3pt plus 1pt minus 1pt}
    \setlength{\belowdisplayskip}{3pt plus 1pt minus 1pt}
    \setlength{\abovedisplayshortskip}{1pt plus 1pt}
    \setlength{\belowdisplayshortskip}{3pt plus 1pt minus 1pt}
}
\usepackage{stfloats}
\usepackage{booktabs}
\usepackage{tabulary}
\usepackage{float}
\makeatletter

\hypersetup{colorlinks=true, linkcolor=black}

\newif\ifapp@intoc
\app@intocfalse

\newcommand{\app@write}[2]{%
  \addtocontents{atoc}{\protect\contentsline{#1}{#2}{\thepage}{\@currentHref}}
}

\let\app@orig@appendix\appendix
\renewcommand{\appendix}{%
  \app@orig@appendix
  \app@intoctrue
}

\let\app@orig@section\section
\renewcommand{\section}{\@ifstar{\app@section@star}{\app@section@nostar}}
\newcommand{\app@section@star}[1]{%
  \app@orig@section*{#1}%
}
\newcommand{\app@section@nostar}{\@ifnextchar[{\app@section@opt}{\app@section@noopt}}
\newcommand{\app@section@opt}[2][]{%
  \app@orig@section[#1]{#2}%
  \ifapp@intoc
    \app@write{section}{\protect\numberline{\thesection}#2}%
  \fi
}
\newcommand{\app@section@noopt}[1]{%
  \app@orig@section{#1}%
  \ifapp@intoc
    \app@write{section}{\protect\numberline{\thesection}#1}%
  \fi
}

\let\app@orig@subsection\subsection
\renewcommand{\subsection}{\@ifstar{\app@subsection@star}{\app@subsection@nostar}}
\newcommand{\app@subsection@star}[1]{%
  \app@orig@subsection*{#1}%
  \ifapp@intoc
    \app@write{subsection}{\protect\numberline{\thesubsection}#1}%
  \fi
}
\newcommand{\app@subsection@nostar}{\@ifnextchar[{\app@subsection@opt}{\app@subsection@noopt}}
\newcommand{\app@subsection@opt}[2][]{%
  \app@orig@subsection[#1]{#2}%
  \ifapp@intoc
    \app@write{subsection}{\protect\numberline{\thesubsection}#2}%
  \fi
}
\newcommand{\app@subsection@noopt}[1]{%
  \app@orig@subsection{#1}%
  \ifapp@intoc
    \app@write{subsection}{\protect\numberline{\thesubsection}#1}%
  \fi
}

\newcommand{\appendixcontentspage}{%
  \clearpage
  \thispagestyle{plain}

  \begin{center}
    {\LARGE \textbf{Appendix}}
  \end{center}

  \vspace{1.2em}
  {\Large \textbf{Contents}\par}
  \vspace{0.8em}

  \begingroup
  \setcounter{tocdepth}{2}

  \renewcommand{\l@section}[2]{%
    \@dottedtocline{1}{0em}{2.2em}{\textcolor{black}{\bfseries ##1}}{\textbf{##2}}%
  }
  \renewcommand{\l@subsection}[2]{%
    \@dottedtocline{2}{2.2em}{3.0em}{\textcolor{black}{##1}}{##2}%
  }

  \@starttoc{atoc}%
  \endgroup

  \clearpage
}

\makeatother

\icmltitlerunning{AffordanceGrasp-R1: Leveraging Reasoning-Based Affordance Segmentation with Reinforcement Learning for Robotic Grasping}

\begin{document}

\twocolumn[
  \icmltitle{AffordanceGrasp-R1: Leveraging Reasoning-Based Affordance Segmentation with Reinforcement Learning for Robotic Grasping}



  \icmlsetsymbol{equal}{*}

  \begin{icmlauthorlist}
    \icmlauthor{Dingyi Zhou}{equal,yyy}
    \icmlauthor{Mu He}{equal,yyy}
    \icmlauthor{Zhuowei Fang}{equal,yyy}
    \icmlauthor{Xiangtong Yao}{yyy}
    \icmlauthor{Yinlong Liu}{sch}
    \icmlauthor{Alois Knoll}{yyy}
    \icmlauthor{Hu Cao}{yyy,comp}
  \end{icmlauthorlist}
  
  \icmlaffiliation{comp}{School of Automation, Southeast University, Nanjing, China}
  \icmlaffiliation{yyy}{Chair of Robotics, Artificial Intelligence and Real-time Systems, Technical University of Munich, Munich, Germany}
  \icmlaffiliation{sch}{Faculty of Data Science, City University of Macau, Macau, China}

  \icmlcorrespondingauthor{Hu Cao}{hu.cao@tum.de}

  \icmlkeywords{Machine Learning, ICML}

  \vskip 0.3in
]



\printAffiliationsAndNotice{}  

\begin{abstract}
Affordance segmentation is crucial for bridging visual perception and robotic manipulation, yet existing Vision–Language Models (VLMs) struggle to align 3D object geometry with the semantic intent of complex grasping instructions due to limited structured reasoning capabilities. We introduce AffordanceGrasp-R1, a reasoning-driven affordance segmentation framework for robotic grasping that combines a chain-of-thought (CoT) cold-start strategy with reinforcement learning to enhance deduction and spatial grounding. In addition, we redesign the grasping pipeline to be more context-aware by generating grasp candidates from the global scene point cloud and subsequently filtering them using instruction-conditioned affordance masks. Extensive experiments demonstrate that AffordanceGrasp-R1 consistently outperforms state-of-the-art (SOTA) methods on benchmark datasets, and real-world robotic grasping evaluations further validate its robustness and generalization under complex language-conditioned manipulation scenarios.
\end{abstract}

\section{Introduction}

Affordance, rooted in ecological psychology~\cite{gibson2014ecological}, characterizes the action possibilities that an object offers to an acting entity. As embodied robots increasingly rely on complex natural language instructions, affordance segmentation has emerged as a critical bridge between high-level semantic intent and low-level motion control. Early research in this area evolved from conventional image-based models~\cite{myers2015affordance, luo2022learning, bahl2023aff, jian2023affordpose, kim2021hotr} to Vision–Language Models (VLMs)~\cite{qian2024affllm, li2024afflear, cuttano2024affclip, chen2025maskprompt, jiao2025freefgrasp, tang2025affordgrasp, yuan2024vlmaff}. Initial VLM-based methods demonstrated that introducing specialized tokens enables language-conditioned pixel-level predictions~\cite{lai2024lisa, ren2024pixellm}. However, these approaches typically rely on static post-training inference and tend to perform shallow semantic matching, failing to capture the deeper relationship between visual-geometric perception and logical reasoning. Effective affordance understanding requires reasoning about how object parts relate to task intent, such as distinguishing between regions that can be safely grasped and those critical for successful task execution~\cite{gibson2014ecological, qian2023understanding}.

\begin{figure}[t!]
  \centering
  \includegraphics[width=\linewidth]{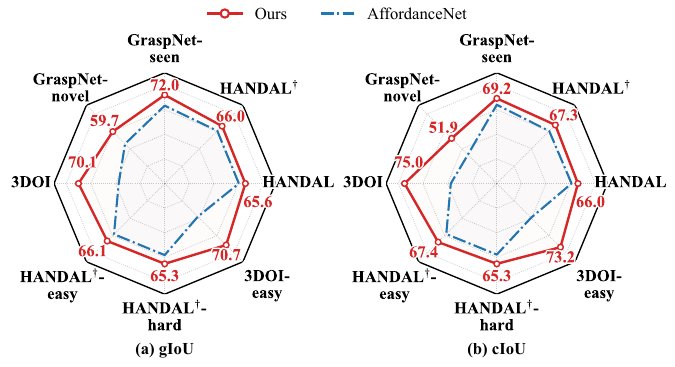}
  \caption{Performance comparison between AffordanceNet~\cite{wu2025ragnet} and our method on grasping datasets, including GraspNet~\cite{fang2020graspnet}, HANDAL~\cite{guo2023handal}, and 3DOI~\cite{qian2023understanding}. }
  \label{fig:radar}
  \vspace{-6mm}
\end{figure}

To address these limitations, we argue that the reasoning capability of Multimodal Large Language Models (MLLMs) must be explicitly enhanced through a structured and robust post-training paradigm. Recent advances, including DeepSeek-R1~\cite{guo2025deepseek} and reasoning-based segmentation methods~\cite{liu2025seg,liu2025visionreasoner}, reveal that relying solely on either supervised fine-tuning (SFT) or reinforcement learning (RL) is insufficient: SFT alone suffers from limited generalization, while RL without proper initialization often exhibits severe training instability. We therefore adopt a synergistic strategy that combines a high-quality Chain-of-Thought (CoT) dataset for cold-start SFT to establish stable reasoning structures, followed by reinforcement learning with carefully designed formatting and accuracy rewards to refine logical consistency through exploration. Meanwhile, we identify a critical mismatch in the downstream transfer from affordance segmentation to grasping. The prevailing approach, AffordanceNet~\cite{wu2025ragnet}, applies the predicted affordance mask to crop the target point cloud before grasp generation. Although this simplifies the input, it discards essential global geometry and contextual information, degrading performance for grasping models pretrained on full-scene point clouds. In contrast, we preserve the global scene point cloud for grasp candidate generation and apply mask-guided filtering afterward, thereby maintaining consistency with the grasp model’s pretraining distribution while improving contextual awareness and grasp robustness.

Building upon these insights, we introduce AffordanceGrasp-R1, a reasoning-driven affordance segmentation framework tailored for robotic grasping. The proposed system enables an action pipeline by translating perceptual inputs into executable grasp guidance through a decoupled and modular architecture. Specifically, our framework comprises three core components: an MLLM module, a segmentation module, and a grasping module. Rather than directly generating pixel-wise masks—which are often noisy and unstable—we guide the MLLM~\cite{bai2025qwen25vl} to produce structured spatial prompts in the form of bounding boxes and points. These prompts are then used by a fine-tuned SAM~2~\cite{ravi2024sam} model to generate high-quality affordance masks. For grasp pose generation, we feed the global scene point cloud into the grasping module and subsequently select the final executable grasp using the predicted affordance mask, thereby preserving global geometric context. The training of AffordanceGrasp-R1 follows a three-stage strategy. First, we perform cold-start alignment of the MLLM using a carefully curated in-domain dataset with high-quality Chain-of-Thought (CoT) supervision to establish stable reasoning behavior. Second, we apply reinforcement learning to further refine logical deduction and spatial reasoning through exploration. Finally, we fine-tune SAM~2 using Low-Rank Adaptation (LoRA)~\cite{hu2022lora} to enable high-fidelity affordance mask generation with minimal disruption to pretrained knowledge.

To demonstrate the effectiveness of our approach, we conduct extensive experiments on the RAGNet benchmark~\cite{wu2025ragnet}. AffordanceGrasp-R1 achieves SOTA performance across all RAGNet sub-datasets, as shown in Fig.~\ref{fig:radar}. In addition, real-world zero-shot robotic grasping experiments yield success rates of 80\% and 72\% under easy and hard language instructions, respectively, highlighting the strong practical grasping capability and robustness of the proposed method.
Our main contributions can be summarized as follows:
\begin{itemize}[nosep]
\item We introduce a high-quality dataset enriched with structured Chain-of-Thought (CoT) reasoning annotations, designed to support effective post-training of MLLMs.
\item We propose a multi-stage post-training framework that synergistically combines SFT and RL with task-oriented reward functions, significantly enhancing the reasoning capability and semantic alignment of MLLMs for complex instruction-conditioned affordance inference.
\item We redesign the grasping pipeline by integrating instruction-conditioned affordance masks with the global scene point cloud during grasp pose selection, improving context awareness and maintaining consistency with the pretraining regime of grasping models. 
\item Our model, AffordanceGrasp-R1, achieves SOTA performance on the RAGNet benchmark and demonstrates strong real-world generalization through zero-shot robotic grasping experiments.
\end{itemize} 
\section{Related Work}

\paragraph{Reasoning-based Segmentation.}
MLLMs are undergoing rapid evolution, transitioning from conventional image captioning and visual question answering towards more complex visual scene understanding tasks that demand stronger logical reasoning, such as image segmentation~\cite{bai2025qwen25vl}~\cite{chen2024far}. The execution of these tasks demands not only highly developed visual perception capabilities but also the capacity for cross-modal semantic reasoning to interpret scene semantics from linguistic intent and to make decisions at the pixel level ~\cite{yang2023lisa++}. Early works, such as LISA~\cite{lai2024lisa}, were notable in their introduction of the concept of special tokens (e.g., \verb|<SEG>|) to establish a connection between the language-reasoning pipeline of MLLMs and the pixel-level prediction capabilities of segmentation models. This seminal research initiated the research direction of reasoning segmentation. Leveraging LISA and subsequent methods, such as OneTokenSegAll~\cite{bai2024one} and PixelLM~\cite{ren2024pixellm}, has demonstrated the feasibility of language-driven pixel segmentation. However, these approaches have also exposed key limitations, namely the requirement for joint fine-tuning of both the MLLM and the segmentation decoder, which imposes substantial demands on hardware and data. Furthermore, due to the absence of fine-grained reasoning annotations in supervised fine-tuning (SFT) datasets, the models demonstrate limited cross-domain generalization, and the generated segmentation masks frequently lack interpretability. To overcome the bottleneck of joint fine-tuning, recent works such as Seg-Zero~\cite{liu2025seg} and SAM-R1~\cite{huang2025sam} have adopted a decoupled design, with the objective of achieving an equilibrium between deployment cost and performance. This paradigm preserves the pixel-level accuracy of existing segmenters while allowing the MLLM to focus solely on reasoning and instruction understanding. With minimal architectural modifications and reduced training overhead, this approach achieves more robust reasoning segmentation, thereby improving practical applicability across diverse domains and real-world scenarios.

\begin{figure}[t!]
    \centering
    \includegraphics[width=0.9\linewidth]{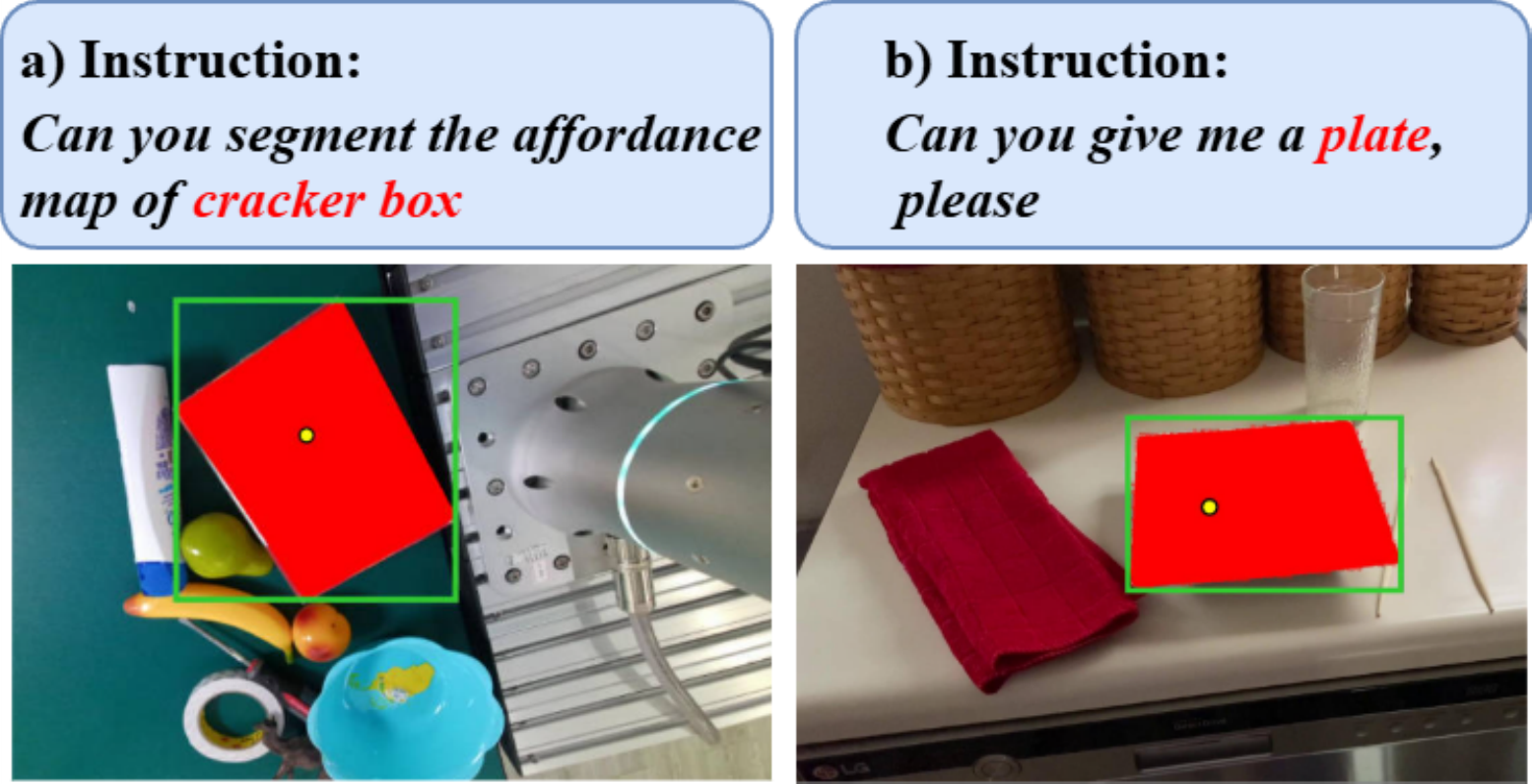}
    \caption{Visualization of the instruction, bounding box, point, and mask annotations. }
    \label{fig:data_visual}
  \vspace{-5mm}
\end{figure}

\paragraph{Affordance Understanding.}
Proposed by James Gibson, affordance provides a shared semantic basis for perception and control in embodied agents~\cite{gibson2014ecological}. Although datasets such as UMD~\cite{myers2015affordance}, OPRA~\cite{fang2018demo2vec}, AGD20k~\cite{luo2022learning}, and HANDAL~\cite{guo2023handal} offer pixel-/region-level supervision, they remain limited in category diversity, scene complexity, and cross-domain generalization~\cite{sawatzky2017weakly}~\cite{luddecke2017learning}. Beyond RGB(-D) perception, neuromorphic/event-based vision has also been investigated for grasp detection and pose estimation under high-speed motion or challenging illumination~\cite{li2020event, cao2022neurograsp}.
RAGNet substantially expands scale and multi-domain coverage and supplies instruction-driven affordance masks~\cite{wu2025ragnet}. Meanwhile, MLLMs excel at instruction following and reasoning~\cite{yang2025qwen3}~\cite{achiam2023gpt} but generally lack fine-grained, affordance-level perception. Relatedly, language-conditioned grasp detection has been explored as a direct means to align linguistic intent with grasp prediction~\cite{jiang2025languageguidedgraspdetectioncoarsetofine}. 
However, these methods typically focus on grasp detection itself and do not explicitly address affordance-centric pixel-level perception or the segmentation-to-3D transfer mismatch in open-world scenes. To bridge the gap between “semantic reasoning” and “pixel-level affordance,” prior work has either linked an affordance VLM with a mask-conditioned grasping network (AffordanceNet)~\cite{wu2025ragnet} or leveraged MLLMs with reinforcement learning to construct an affordance localization framework (Affordance-R1)~\cite{wang2025affordance}, yet an MLLM-based affordance perception solution with combined SFT and RL strategies tailored to robotic grasping is still lacking. Recent trends indicate that higher-level reasoning signals (such as CoT) and post-training methods (such as the combination of SFT and RL) are being introduced into multimodal accessibility comprehension tasks. The stepwise reasoning structure of CoT assists models in capturing object functions and scene semantics, while the joint SFT+RL post-training paradigm enhances models' stable learning of task logic and their generalization capabilities across complex scenarios. However, within the domain of affordance understanding, these approaches have yet to be systematically integrated.

\begin{figure}[t!]
    \centering
    \includegraphics[width=\linewidth]{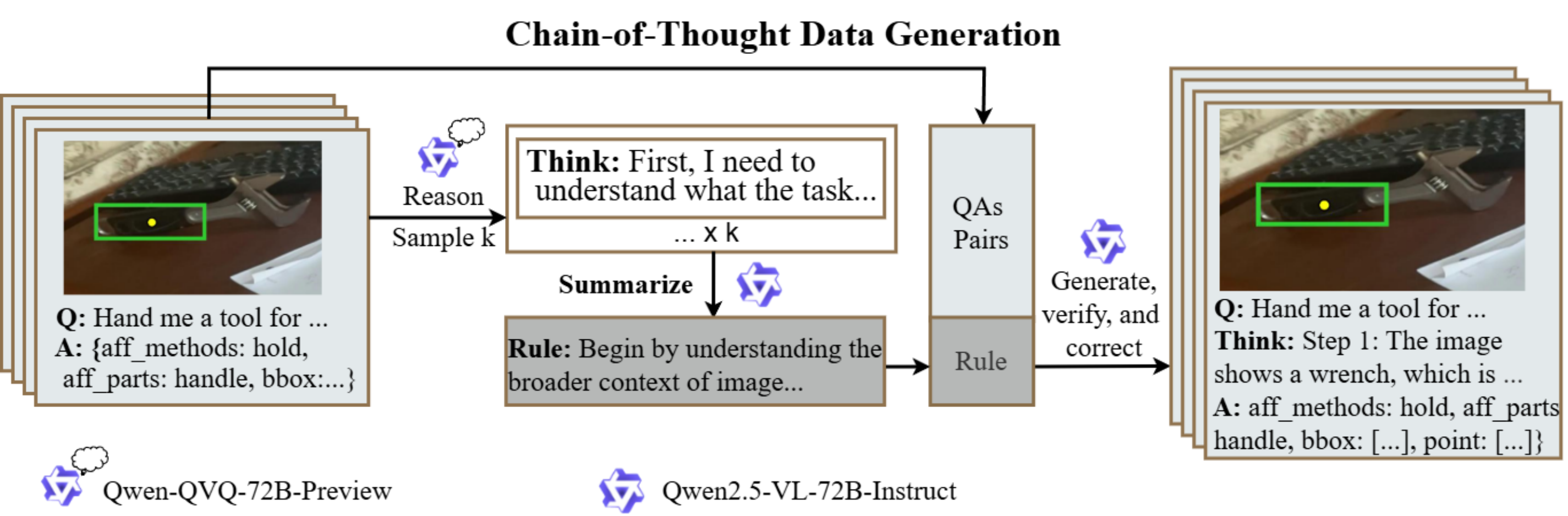}
    \caption{
          Overview of the CoT data generation process. 
    }
    \label{fig:cot_affordance_example}
  \vspace{-5mm}
\end{figure}

\section{Dataset}
\label{sec:dataset}
Building upon the RAGNet~\cite{wu2025ragnet} dataset, we construct a sampled dataset for post-training the proposed model. This dataset comprises 42,400 samples, among which 3,313 are allocated for CoT cold-start training and the remaining 39,087 are used for RL post-training.

\paragraph{Box-point prompt generation.}
Based on the original mask annotations, we extract the leftmost, topmost, rightmost, and bottommost pixels to construct the bounding box $B = [B_{x1}, B_{y1}, B_{x2}, B_{y2}]$. Additionally, we compute the center point of the largest inscribed circle within the mask, denoted as $P = (P_x, P_y)$. Representative examples are shown in Fig.~\ref{fig:data_visual}. To ensure that the extracted prompts ($B$ and $P$) can reliably reconstruct the original affordance mask, we use a self-consistency verification mechanism. Specifically, the box-point prompt $(B,P)$ is provided to SAM 2~\cite{ravi2024sam} to generate a predicted mask $\hat{M}$. The Intersection over Union (IoU) is then computed between the predicted mask $\hat{M}$ and the original RAGNet mask annotation $M$. This process can be formulated as:
\begin{equation}
    \mathrm{IoU}(M, \hat{M}) = \frac{|M \cap \hat{M}|}{|M \cup \hat{M}|}.
\end{equation}
Specifically, samples with $\text{IoU} < 0.6$ are discarded to ensure the accuracy and consistency of the generated box–point prompts.

\paragraph{CoT data generation.}
We first generate two semantic labels: \textit{aff\_methods} and \textit{aff\_parts}. Specifically, \textit{aff\_methods} denotes the functional affordance type associated with an object (e.g., grasp, pull, hold), while \textit{aff\_parts} specifies the object regions that enable the affordance (e.g., handle, whole). These two labels constitute part of the answer.
As illustrated in Fig.~\ref{fig:cot_affordance_example}, the pre-trained visual reasoning model Qwen-QVQ-72B-Preview generates reasoning trajectories from a sampled set of k input images and QA pairs. These reasoning processes are subsequently summarized and distilled into generalizable reasoning rules using Qwen2.5-VL-72B-Instruct. By providing these rules together with the original QA pairs to Qwen2.5-VL-72B-Instruct, we generate new QA pairs annotated with CoTs.
Finally, we employ Qwen2.5-VL-72B-Instruct to evaluate each generated CoT along three dimensions: logical consistency, reliance on visual evidence, and completeness of the inference steps. Samples that fail to satisfy the criterion are automatically corrected or regenerated to ensure the quality and correctness of the resulting CoTs. The prompts for generating CoT reasoning are presented in the Appendix~\ref{sec:cot_prompt}.

\begin{figure*}[t!]
   \centering
   \includegraphics[width=0.95\linewidth]{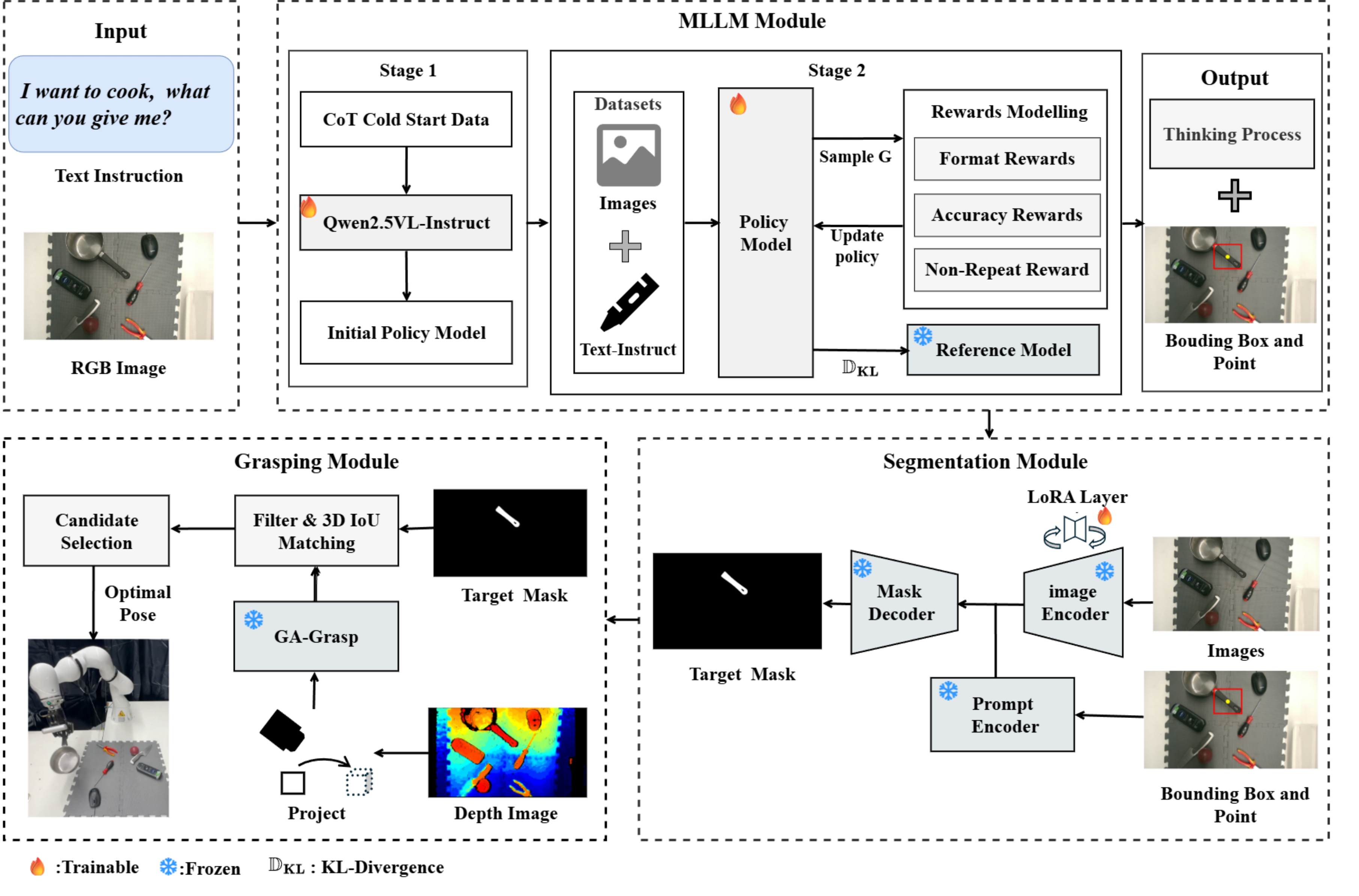}
   \caption{Overview of AffordanceGrasp-R1. A Qwen2.5-VL-7B model with two-stage post-training serves as the MLLM module, while SAM~2 with LoRA fine-tuning performs affordance segmentation. The resulting mask guides object point cloud selection, and the optimal grasp pose is chosen from candidate grasps.}
   \label{fig:fig_method}
\end{figure*}

\section{Method}
\label{sec:method}

As shown in Fig.~\ref{fig:fig_method}, we propose AffordanceGrasp-R1, which consists of an MLLM module, a segmentation module, and a grasping module. Specifically, the Qwen2.5-VL-7B model with two-stage post-training serves as the MLLM module, SAM~2 with LoRA fine-tuning is employed as the segmentation module, and the pre-trained GA-Grasp is used as the grasping module.

\subsection{Multimodal Large Language Model (MLLM)}
We employ  Qwen2.5-VL-7B~\cite{bai2025qwen25vl} as the MLLM $F$. Given an image–instruction pair $(I,T)$, the model performs multimodal reasoning to generate intermediate thinking outputs along with structured spatial prompts, including a bounding box $B$ and a point $P$. This process is formalized as follows:
\begin{equation}
{(B, P)}=F(I,T).
\end{equation}
Specifically, MLLM includes two post-training stages: CoT cold start and reforcement learning (RL).
\paragraph{Stage 1: CoT Cold Start.}
\label{sec:sft}
Although recent works~\cite{liu2025seg, liu2025visionreasoner} have shown that training MLLMs from scratch using pure reinforcement learning (RL) is theoretically feasible, directly initializing RL from a base model often suffers from the cold-start problem in practice. Specifically, the initial policy lacks structured reasoning capabilities, resulting in inefficient exploration, unstable training dynamics, and a tendency toward suboptimal convergence~\cite{guo2025deepseek}. To address this issue, we first construct a carefully curated cold-start dataset (see Sec.~\ref{sec:dataset}) and perform supervised fine-tuning (SFT) on the Qwen-2.5VL-7B-Instruct model to endow it with stable reasoning behaviors. Specifically, each training sample is represented as $(I, T, r, a)$, where $r=(r_1,\ldots,r_{T_r})$ corresponds to the reasoning steps, and $a=(a_1,\ldots,a_{T_a})$ is the final answer, with ${T_r}$ and ${T_a}$ denoting the token-level sequence lengths of reasoning steps $r$ and 
the final answer $a$, respectively. The model is trained autoregressively to maximize the likelihood of generating $r$ and 
$a$ conditioned on image-instruction pair$(I,T)$. The supervised fine-tuning objective is defined as follows:
\begin{align}
\mathcal{L}_{\text{SFT}}(\theta)
= -\mathbb{E}_{(I,T,r,a) \sim \mathcal{D}} \Bigg[
& \sum_{t=1}^{T_r} \log \pi_\theta(r_t \mid I, T, r_{<t}) \nonumber\\
& + \sum_{t=1}^{T_a} \log \pi_\theta(a_t \mid I, T, r, a_{<t})
\Bigg].
\label{eq:sft-loss}
\end{align}
where $\mathcal{D}_{CoT}$ denotes the CoT training dataset, and $\pi_\theta$ denotes the token probability from the model. By minimizing the SFT objective, we obtain the initial policy for RL:
\begin{equation}
\pi_{\mathrm{CoT}} = \arg\min_{\theta} \mathcal{L}_{\mathrm{SFT}}(\theta).
\end{equation}
By providing a stable initialization, $\pi_{\mathrm{CoT}}$ lays a solid foundation for subsequent visual reasoning based on reinforcement learning.

\paragraph{Stage 2: Reinforcement Learning.}
\label{sec:grpo}
Group Relative Policy Optimization (GRPO)~\cite{shao2024deepseekmath} is a critic-free RLHF (Reinforcement Learning from Human Feedback) / RLVR (Reinforcement Learning with Verifiable Rewards) algorithm, subsequently adopted in the DeepSeek R1-Zero line. GRPO replaces PPO’s~\cite{schulman2017proximal} reliance on a learned value function with in-group comparison. Given the prompt $q$, the old policy $\pi_{\text{old}}$ samples $G$ candidate outputs $\{o_i\}$. Each candidate is evaluated by a task-specific reward function $R(q,o_i)$, producing rewards $R_i$. The rewards are then normalized within the group to compute the corresponding advantages:
\begin{equation}
{A}_{i}
=
\frac{r_i - \mathrm{mean}\!\left(\{R_i\}_{i=1}^{G}\right)}
{\mathrm{std}\!\left(\{R_i\}_{i=1}^{G}\right)}.
\end{equation}
where $\mathrm{mean}\!\left(\{R_i\}_{i=1}^{G}\right)$ and $\mathrm{std}\!\left(\{R_i\}_{i=1}^{G}\right)$ denote the mean and standard deviation of the rewards within the group, respectively. GRPO then maximizes a clipped objective augmented with KL penalty to constrain policy updates:
\begin{align}
\mathcal{J}_{\mathrm{GRPO}}(\theta)
= {} & \mathbb{E}_{q}\,
       \mathbb{E}_{\{o_i\}\sim\pi_{\mathrm{old}}(\cdot\mid q)}
       \Bigg[
       \frac{1}{G}\sum_{i=1}^{G} \min\{ s_1 A_i,\; s_2 A_i \}
       \nonumber\\
    & 
       -\; \beta\, D_{\mathrm{KL}}(\pi_{\theta}\,\|\,\pi_{\mathrm{CoT}})
       \Bigg],
\label{eq:grpo-obj}
\end{align}
\begin{equation}
s_1=\frac{\pi_\theta(o_i|q)}{\pi_{\text{old}}(o_i|q)},
\qquad
s_2=\mathrm{clip}\!\left(s_1,\,1-\varepsilon,\,1+\varepsilon\right).
\label{eq:grpo-ratios}
\end{equation}
where the KL divergence is defined as:
\begin{equation}
D_{\mathrm{KL}}\!\left(\pi_\theta \,\|\, \pi_{\mathrm{CoT}}\right)
= \frac{\pi_{\mathrm{CoT}}(o_i\mid q)}{\pi_\theta(o_i\mid q)}
 - \log \frac{\pi_{\mathrm{CoT}}(o_i\mid q)}{\pi_\theta(o_i\mid q)} - 1 .
\label{eq:kl}
\end{equation}

\begin{figure*}[t!]
  \centering
  \includegraphics[width=0.8\linewidth]{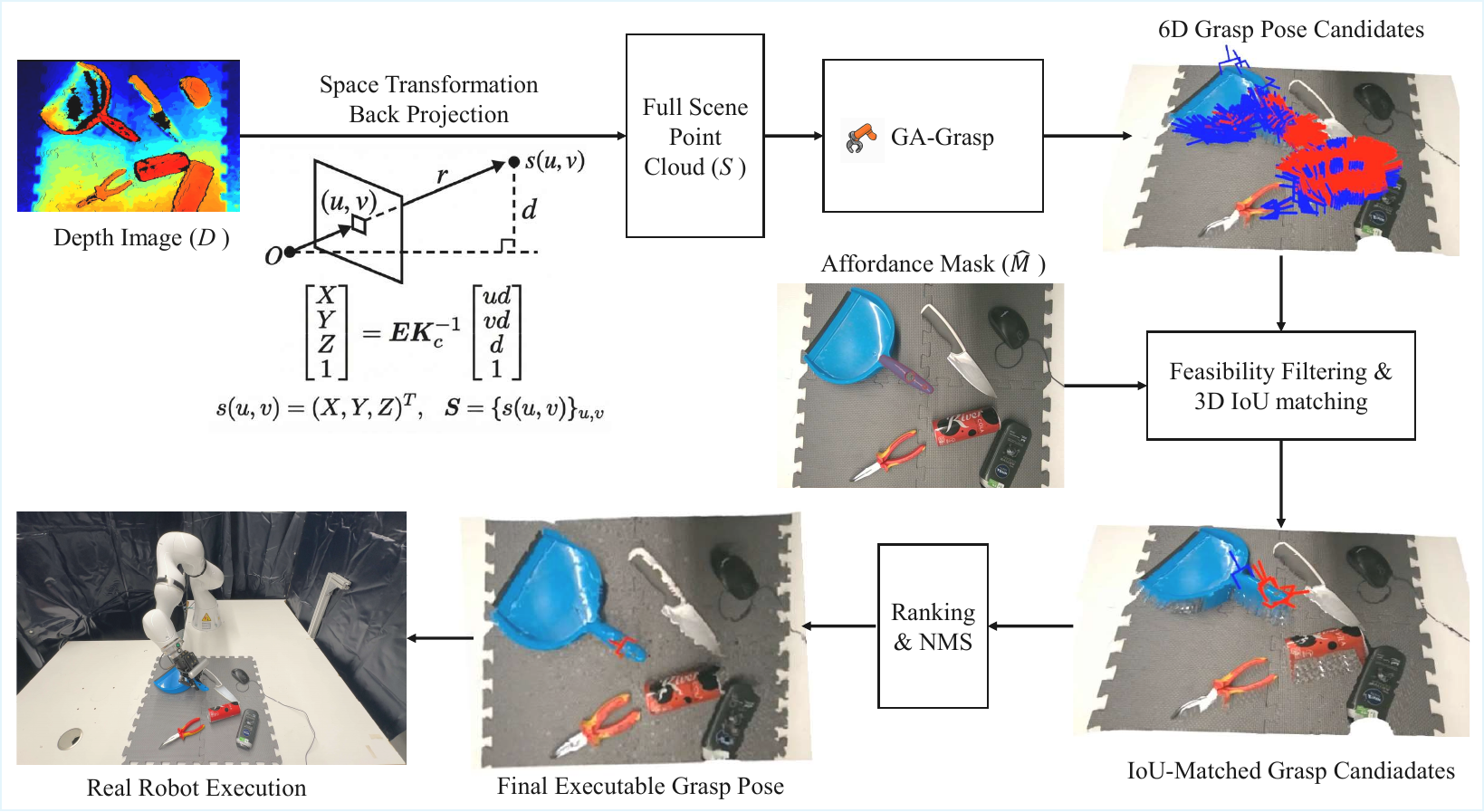}
  \caption{Overview of grasp pose generation. 6D grasp candidates are generated from the full-scene point cloud and then refined through mask-guided semantic filtering, followed by score ranking and NMS, to obtain the final executable grasp pose.}
  \label{fig:grasp_pose_generation}
\end{figure*}

By deriving a baseline from group-level statistics, GRPO eliminates the need to train a separate critic. The clipping mechanism stabilizes policy updates, while the KL regularization term mitigates reward hacking. In practice, the reward function typically combines accuracy and rule- or format-compliance components with appropriate normalization and weighting, enabling robust reasoning performance under simplified supervision.

For RL training, we design a composite reward consisting of a \textbf{format reward}, a \textbf{non-repeat reward}, and an \textbf{accuracy reward}. \textbf{Format reward} enforces parsable and executable structured outputs. It comprises two components: 1) \textbf{Thinking format reward}, which constrains the model to place its reasoning process within \texttt{<think> … </think>} tags and to produce the final answer only afterward, explicitly separating thinking from answering. The final answer must be enclosed within \texttt{<answer> … </answer>} tags. 2) \textbf{Answer format reward}, which requires the final answer to follow a fixed schema for downstream parsing and prompt construction, e.g.,
\{bbox:[x1,y1,x2,y2], point:[x,y], aff\_methods: grasp, aff\_parts: handle (whole)\}. Only the predefined keys are accepted, and all coordinates are verified to lie within the image bounds. Outputs passing these checks receive a positive reward. \textbf{Non-repeat reward} penalizes templated or repetitive reasoning. Specifically, the content within the \texttt{<think>} block is segmented into sentence-level units, and repeated patterns are penalized, while non-redundant, image- and instruction-grounded reasoning is rewarded. This encourages a think-then-answer behavior and reduces hallucinations caused by mechanical phrasing. \textbf{ Accuracy reward} evaluates spatial localization and geometric precision via one-to-one matching. For each satisfied criterion, an increment of $\frac{1}{\max\{N,K\}}$ is added. $N$ is the number of ground-truth boxes and $K$ denotes the number of predicted boxes. The criteria include: 1) \textbf{Bounding box IoU reward}: one-to-one IoU is computed between predicted and ground-truth boxes; If IoU $> 0.5$, the increment is added. 2) \textbf{Bounding box L1 reward}: the L1 distance over $(x_{\min},y_{\min},x_{\max},y_{\max})$ is computed for matched boxes; if the distance is below \textbf{10 pixels}, the increment is added. 3) \textbf{Keypoint L1 reward}: the L1 distance is computed for both primary and auxiliary keypoints; we add corresponding increment if the distance for a keypoint is below \textbf{30 pixels}.

\subsection{Segmentation Module}
The segmentation module $O$ adopts SAM 2~\cite{ravi2024sam}. Given the original image $I$ and the prompt $(B,P)$ generated by the MLLM, $O$ produces the final affordance mask. We jointly leverage $B$ and $P$ to balance global constraints and local detail: the bounding box restricts the candidate region, while the point provides positive guidance and disambiguation within the box. This process can be formulated as follows:
\begin{equation}
M = O(I;B,P).
\end{equation}
Since the pretrained distribution of general segmentation in SAM 2 differs from that of affordance segmentation, directly using the pretrained SAM 2 results in suboptimal performance. To obtain a model better aligned with our task, we fine-tune SAM 2 using the generated prompts $(B_i, P_i)$ and the corresponding ground-truth masks $M_i$. 
Specifically, we adopt Low-Rank Adaptation (LoRA)~\cite{hu2022lora} to achieve effective in-domain adaptation while largely preserving the pretrained knowledge. LoRA introduces low-rank updates to the projection matrices within the attention mechanism (i.e., the query, key, and value matrices), updating the weights as follows:
\begin{equation}
    h = W_0x + \Delta Wx = W_0x+\Phi_\text{up}\Phi_\text{down}^Tx,
\end{equation}
where $W_0\in\mathbb{R}^{d\times k}$ indicates the pretrained weights that project the features from $d$ dimension to $k$ accordingly. $\Delta W$ is the adaptation that we estimate with the projection matrix with $\Phi_\text{up}\in\mathbb{R}^{d\times r}$ and $,\Phi_\text{down}\in\mathbb{R}^{r\times d}$, where the rank $r \ll min(d,k)$. 
\subsection{Grasping Module}
\label{sec:pose}
Many prior works~\cite{cao2021residual, cao2023effgrasp} have explored real-time planar grasp detection from RGB-D inputs. In our framework, the RGB image is first fed into the MLLM to generate bounding box and point prompts, which guide SAM 2 to produce the final affordance mask $\hat{M}\!\in\!\{0,1\}^{H\times W}$. In parallel, we perform a space transformation on the depth input. Each valid pixel $(u,v)$ with depth $d=D(u,v)$ is back-projected using the camera intrinsics $K_c\!\in\!\mathbb{R}^{3\times3}$ and extrinsics $E=\begin{bmatrix}R & t\\ \mathbf{0}^\top & 1\end{bmatrix}\!\in\!\mathbb{R}^{4\times4}$ to construct an ordered scene point cloud,
\begin{equation}
\begin{bmatrix}X\\Y\\Z\\1\end{bmatrix}
= E\,K_c^{-1}
\begin{bmatrix}u\,d\\ v\,d\\ d\\ 1\end{bmatrix},
\end{equation}
\begin{equation}
s(u,v)=(X,Y,Z)^\top,\quad
\mathcal{S}=\{s(u,v)\}_{u,v}.
\label{eq:backproj}
\end{equation}
where, $d$ denotes the depth along the camera optical axis (orthogonal to the image plane), and $(X,Y,Z)$ are the corresponding 3D coordinates in the world coordinate frame. 

As illustrated in Fig.~\ref{fig:grasp_pose_generation}, the full-scene point cloud $P$ is fed into the grasping model GA-Grasp~\cite{cao2025geometryaware} to generate 6D grasp candidates. Grasp candidates are generated from the entire scene (i.e., across all visible objects), which preserves global geometric context and improves robustness to occlusion and depth noise. Finally, we fuse RGB semantics with depth geometry by retaining only those 3D points whose corresponding pixels lie within the affordance region, producing a semantics-aligned subcloud that is used to filter the grasp candidates:
\begin{equation}
\Omega=\{(u,v)\mid \hat{M}(u,v)=1\},
\end{equation}
\begin{equation}
\hat S=\{\,s(u,v)\mid (u,v)\in\Omega\,\}\ \equiv\ S\otimes \hat{M} .
\label{eq:maskfusion}
\end{equation}
\paragraph{Comparison to AffordanceNet~\cite{wu2025ragnet}.}
A key distinction from AffordanceNet lies in the stage at which the predicted affordance mask is applied. AffordanceNet applies the mask before grasp generation to obtain a target-aligned point cloud, which is then used as input to the grasping model to generate grasp candidates. In contrast, we first generate grasp candidates from the full-scene point cloud $S$ and subsequently perform mask-guided semantic filtering using the semantics-aligned subcloud $\hat S$. After obtaining a set of 6D grasp candidates, we apply a 3D IoU matching to identify grasps that best align with the target affordance region. This strategy improves target grasp selection accuracy, as the grasping model is pretrained to operate on full-scene point clouds rather than on pre-filtered target regions. Specifically, we first derive a semantics-aligned subcloud $\hat S$, which serves as the target reference in 3D space and defines the target volume for IoU matching. As illustrated in Appendix~\ref{sec:3D_IoU_Matching}   (Fig.~\ref{fig:3d_iou_matching}), for each generated grasp candidate, we compute a 3D IoU score between the gripper closing volume and the semantics-aligned target region induced by $\hat S$. Intuitively, a high-IoU grasp exhibits substantial overlap with $\hat S$, indicating strong semantic alignment, whereas a low-IoU grasp shows limited intersection. This IoU-based metric quantifies the degree of alignment and enables effective prioritization of grasp candidates. Finally, candidates with the highest IoU scores are selected optionally via thresholding or Top-$K$ selection ensuring that the chosen grasps are both physically feasible and well aligned with the intended target object.
In practice, we first discard geometrically infeasible candidates using routine checks, such as collision detection against the full-scene cloud $S$ and basic kinematic or gripper-width constraints. For the remaining candidates, we compute a 3D IoU score between each gripper closing/contact volume and the target volume defined by the semantics-aligned subcloud $\hat{S}$, retaining the Top-$K$ candidates with the highest IoU scores. These Top-$K$ candidates are then ranked by grasp confidence, and non-maximum suppression (NMS) is applied, resulting in a grasp pose for real-world execution.

\begin{table*}[t!]
\centering
\caption{Affordance segmentation on the main subsets (gIoU / cIoU, \%). Best results are in bold.}
\resizebox{\linewidth}{!}{
\begin{tabular}{lcccccc}
\toprule
\textbf{Model} &
\textbf{HANDAL} &
\textbf{HANDAL$^\dagger$} &
\textbf{GraspNet-seen} &
\textbf{GraspNet-novel} &
\textbf{3DOI} &
\textbf{Average}\\
& (gIoU / cIoU)\% & (gIoU / cIoU)\% & (gIoU / cIoU)\% & (gIoU / cIoU)\% & (gIoU / cIoU)\% & (gIoU / cIoU)\%\\
\midrule
VLPart~\cite{sun2023going} + SAM 2~\cite{ravi2024sam}           & 40.9 / 28.9 & 40.7 / 27.6 & --          & --          & --   & --        \\
GroundingDINO~\cite{liu2024grounding} + SAM 2~\cite{ravi2024sam}    & 34.7 / 26.8 & 34.9 / 26.9 & --          & --          & --     & --     \\
Florence 2~\cite{xiao2024florence} + SAM 2~\cite{ravi2024sam}        & 39.7 / 22.4 & 39.4 / 22.5 & --          & --          & --     & --     \\
LISA-7B~\cite{lai2024lisa}                      & 16.2 / 12.0 & 15.4 / 11.8 & 17.7 / 17.7 & 25.2 / 24.1 & 21.5 / 13.7 & 19.2 / 15.9\\
GLaMM~\cite{rasheed2024glamm}                       & 24.9 / 17.2 & 25.1 / 17.0 & 21.6 / 10.5 & 19.2 / 8.6  & 19.7 / 14.1  & 22.1 / 13.5\\
Qwen2.5VL-7B~\cite{bai2025qwen25vl} + SAM 2~\cite{ravi2024sam}        & 8.4 / 13.6          & 20.6 / 17.0 & 30.1 / 29.6 & 38.7 / 37.1 & 31.5 / 32.3 & 25.9 / 25.9\\
Affordance-R1~\cite{wang2025affordance}                       & 27.5 / 20.4 & 17.6 / 10.3 & 50.8 / 46.3 & 61.3 / 53.3 & 51.0 / 40.9  & 41.6 / 34.2\\
Segzero~\cite{liu2025seg}                      & 29.5 / 21.0          & 30.3 / 21.4 & 51.8 / 43.3 & 59.2 / 49.6 & 50.2 / 30.0 &   44.2 / 33.1\\
Visionreasoner~\cite{liu2025visionreasoner}              & 27.6 / 18.8          & 28.3 / 18.3 & 46.4 / 36.5 & 51.9 / 40.4 & 49.0 / 29.4 & 40.6 / 28.7 \\
AffordanceNet~\cite{wu2025ragnet}                       & 60.3 / 60.8 & 60.5 / 60.3 & 63.3 / 64.0 & 45.6 / 33.2 & 37.4 / 37.4  & 53.4 / 51.1\\
\midrule
\textbf{AffordanceGrasp-R1}  & \textbf{65.6} / \textbf{66.0}          & \textbf{66.0} / \textbf{67.3} & \textbf{72.0} / \textbf{69.2} & \textbf{59.7} / \textbf{51.9} & \textbf{70.1} / \textbf{75.0} & \textbf{66.7} / \textbf{65.9} \\
\bottomrule
\end{tabular}
}
\label{tab:affseg_main_only}
\end{table*}

\section{Experiments on Visual Affordance}

\subsection{ Results}
\label{sec:results}

\paragraph{Evaluation on Affordance Segmentation.} The model is evaluated on affordance segmentation benchmarks~\cite{wu2025ragnet}, including HANDAL~\cite{guo2023handal}, GraspNet~\cite{fang2020graspnet}, and 3DOI~\cite{qian2023understanding}. As shown in Tab.~\ref{tab:affseg_main_only}, AffordanceGrasp-R1 achieves SOTA performance across all evaluated datasets, demonstrating consistent superiority in both in-domain and out-of-domain (OOD) settings. Appendix~\ref{sec:experiment_results}  (Fig.~\ref{fig:affordance_visual_results}) further provides a qualitative comparison between AffordanceGrasp-R1 and AffordanceNet~\cite{wu2025ragnet} on affordance segmentation tasks. Notably, AffordanceGrasp-R1 exhibits particularly strong performance in challenging scenes that require high-precision functional segmentation.

\paragraph{Evaluation on Reasoning Affordance.}
Tab.~\ref{tab:affseg_reason_only} presents results on the reasoning-based subsets of HANDAL$^\dagger$  and 3DOI, which specifically evaluate reasoning-driven affordance segmentation. Our method achieves SOTA performance across all settings, substantially outperforming all competing approaches. On the 3DOI dataset, AffordanceGrasp-R1 surpasses AffordanceNet by 32.6\% gIoU points and 33.8\% cIoU points. Similar advantages are observed on both HANDAL-easy and HANDAL-hard, where our model consistently attains the highest scores. Additional visualizations and qualitative results are provided in Appendix~\ref{sec:experiment_results} (Fig.~\ref{fig:reasoning__visual_results}). We attribute these gains to the synergistic combination of CoT-based cold start initialization and reinforcement learning (RL). The CoT initialization establishes stable reasoning structures, while GRPO RL further refines the affordance reasoning policy, resulting in enhanced reasoning capability.

\begin{table*}[t!]
\centering
\caption{Affordance segmentation on the reasoning-based subsets (gIoU / cIoU, \%). Best results are in bold.}
\resizebox{\linewidth}{!}{
\begin{tabular}{lcccc}
\toprule
\textbf{Model} &
\textbf{HANDAL$^\dagger$-easy} &
\textbf{HANDAL$^\dagger$-hard} &
\textbf{3DOI} &
\textbf{Average}\\
& (gIoU / cIoU)\% & (gIoU / cIoU)\% & (gIoU / cIoU)\% & (gIoU / cIoU)\%\\
\midrule

GroundingDINO~\cite{liu2024grounding} + SAM 2~\cite{ravi2024sam}    & 3.6 / 3.0   & 3.4 / 3.1   & 4.1 / 3.9  & 3.7 / 2\\
LISA-7B~\cite{lai2024lisa}                      & 15.5 / 11.9 & 12.3 / 8.1  & 12.3 / 8.1  & 3.7 / 3.3\\
GLaMM~\cite{rasheed2024glamm}                  & 4.7 / 3.5   & 5.0 / 3.5   & 4.4 / 2.9  & 4.7 / 3.3\\
Qwen2.5VL-7B~\cite{bai2025qwen25vl} + SAM 2~\cite{ravi2024sam}  & 20.6 / 18.1 & 21.9 / 17.2 & 38.4 / 30.8  &  27.0 / 22.0\\
Affordance-R1~\cite{wang2025affordance}    & 19.4 / 11.7 & 15.9 / 8.5 & 50.1 / 41.5 & 28.5 / 20.6\\
Segzero~\cite{liu2025seg}                      & 29.8 / 20.6 & 27.6 / 16.8 & 49.3 / 29.4 & 35.6 / 22.3\\
Visionreasoner~\cite{liu2025visionreasoner}    & 29.4 / 20.4 & 27.1 / 15.8 & 50.3 / 30.8 & 35.6 / 22.3\\
AffordanceNet~\cite{wu2025ragnet}                     & 58.3 / 58.1 & 58.2 / 57.8 & 38.1 / 39.4 & 51.5 / 51.8\\
\midrule

\textbf{AffordanceGrasp-R1} & \textbf{66.1} / \textbf{67.4} & \textbf{65.3} / \textbf{66.4} & \textbf{70.7} / \textbf{73.2} & \textbf{67.4} / \textbf{69.0} \\
\bottomrule
\end{tabular}
}
\label{tab:affseg_reason_only}
\end{table*}

\begin{table*}[t!]
    \centering
    \caption{Comparison of real-world grasping success rates (\%) between AffordanceGrasp-R1 and AffordanceNet under easy and hard reasoning instructions. Each task is executed ten times, and the average success rate is reported.}
    \label{tab:real_robot_comparison}
    \resizebox{\linewidth}{!}{
    \begin{tabular}{llccccccccccc}
        \toprule
        Method & Instruction & Banana & Can & Screwdriver & Hammer & Pot & Mouse & Ball & Scoop & Knife & Glove & Average \\
        \midrule
        \multirow{2}{*}{AffordanceNet~\cite{wu2025ragnet}}
          & Easy  & 70 & 70 & 50 & 70 & 60 & 60 & 60 & 60 & 50 & 70 & 62 \\
          & Hard  & 60 & 60 & 30 & 60 & 50 & 40 & 40 & 50 & 50 & 60 & 50 \\
        \midrule
        \multirow{2}{*}{AffordanceGrasp-R1}
          & Easy  & \textbf{90} & \textbf{80} & \textbf{80} & \textbf{80} & \textbf{80} & \textbf{80} & \textbf{70} & \textbf{90} & \textbf{70} & \textbf{90} & \textbf{80} \\
          & Hard  & \textbf{90} & \textbf{70} & \textbf{70} & \textbf{70} & \textbf{80} & \textbf{60} & \textbf{70} & \textbf{70} & \textbf{60} & \textbf{80} & \textbf{72} \\
        \bottomrule
    \end{tabular}
    }
\end{table*}

\paragraph{Evaluation on Real Robot.}
\label{sec:main_results_real_robot}
Tab.~\ref{tab:real_robot_comparison} summarizes the real-world grasp success rates of AffordanceGrasp-R1 and AffordanceNet under easy and hard reasoning instructions in a strict zero-shot setting. Overall, AffordanceGrasp-R1 consistently outperforms AffordanceNet across all ten objects. Specifically, the average success rate increases from 62\% to 80\% under easy instructions and from 50\% to 72\% under hard instructions. These results indicate that our method performs reliably not only under straightforward language prompts but also under more demanding semantic constraints that require precise part-level grounding. A closer inspection reveals that the performance gains are most pronounced for objects requiring fine-grained localization of actionable regions, particularly tool-like categories with small handles or narrow graspable parts. For instance, on the screwdriver task, AffordanceNet achieves success rates of 50\% (easy) and 30\% (hard), whereas AffordanceGrasp-R1 improves these results to 80\% (easy) and 70\% (hard). A similar trend is observed for the pot category: AffordanceNet drops from 60\% (easy) to 50\% (hard), while AffordanceGrasp-R1 maintains high performance at 80\% (easy and hard). These comparisons suggest that our approach better preserves grasp reliability when instructions demand precise localization of the intended actionable part. Moreover, the reduced performance gap between easy and hard instructions indicates that AffordanceGrasp-R1 is more robust to reasoning-intensive language constraints. We attribute these gains to our reasoning-driven affordance segmentation, which produces more accurate and reliable instruction-conditioned target regions for subsequent grasp candidate filtering, thereby mitigating failures caused by grasping non-actionable parts. We observe two dominant failure modes: (i) slight misalignment in the predicted affordance region, which can suppress valid grasp candidates during filtering or ranking; and (ii) grasps whose contact points fall on non-actionable parts (e.g., the head of a hammer or unstable edges), leading to slippage or unstable pickups. Representative instruction–image pairs are shown in Appendix~\ref{sec:experiment_results} (Fig.~\ref{fig:easy_hard_examples}), while qualitative real-world rollout results are provided in Appendix~\ref{sec:experiment_results} (Fig.~\ref{fig:rollout_affordance} and Fig.~\ref{fig:rollout_direct}).

\begin{table}[t!]
    \centering
    \caption{Ablation study on the effectiveness of each component. }
    \label{tab:ablation}
    \resizebox{\linewidth}{!}{
    \begin{tabular}{l c c c}
        \toprule
        \textbf{Configuration} &
        \textbf{Main} &
        \textbf{Reasoning} &
        \textbf{All} \\
        & (gIoU / cIoU)\% & (gIoU / cIoU)\% & (gIoU / cIoU)\% \\
        \midrule
        Qwen2.5VL-7B + SAM 2& 25.9 / 25.9 & 27.0 / 22.0 & 28.8 / 26.0 \\
        \midrule
        SFT & 49.5 / 37.9 & 51.7 / 39.6 & 52.0 / 40.3 \\
        CoT-SFT  & 55.0 / 46.9 & 52.3 / 42.9 & 54.0 / 45.4 \\
        CoT-SFT + SFT  & 65.2 / 64.2 & 64.5 / 64.3 & 64.9 / 64.3 \\
        CoT-SFT + GRPO & 65.5 / 65.0 & 65.1 / 67.3 & 65.4 / 65.9 \\
        CoT-SFT + GRPO + SAM 2-LoRA & \textbf{66.7} / \textbf{65.9} & \textbf{67.4} / \textbf{69.0} & \textbf{66.9} / \textbf{67.1} \\
        \bottomrule
    \end{tabular}
    }
    \label{tab:training effects}
\end{table}
\subsection{Ablation Study}
\label{sec:Ablation Study in experiment on Visual Affordance-Grasping}

Tab.~\ref{tab:training effects} presents the impact of each training strategy on overall performance. Replacing standard SFT with CoT-SFT led to a notable improvement, with the average gIoU/cIoU across all subsets increasing from 52.0\% / 40.3\% to 54.0\% / 45.4\%. These results indicate that explicit supervision of the reasoning process enhances segmentation quality in complex semantic scenarios. The subsequent large-scale training stage further boosts the performance. Notably, GRPO achieves a more stable cIoU improvement on reasoning subsets, suggesting that it effectively translates refined reasoning into accurate perceptual grounding. Finally, after incorporating SAM2-LoRA fine-tuning, the average cIoU across all subsets reached 67.1\%, with reasoning subsets achieving 69.0\%. This shows that lightweight adaptation of the segmentation module converts MLLM reasoning capabilities into more precise pixel-level affordance predictions.


\section{Conclusion}

In this work, we introduced AffordanceGrasp-R1, a reasoning-driven affordance segmentation framework that effectively bridges language understanding, visual perception, and robotic grasping. By combining a high-quality CoT cold-start dataset with a stage-wise post-training strategy based on GRPO, our approach significantly improves reasoning and spatial grounding under complex language instructions. Coupled with prompt-guided SAM 2 for fine-grained affordance segmentation and a global point-cloud–mask filtering strategy for context-aware grasp pose selection, AffordanceGrasp-R1 achieves SOTA performance across benchmark datasets and demonstrates strong zero-shot generalization in real-world robotic grasping tasks. These results highlight the importance of structured reasoning for affordance understanding and suggest promising directions for extending the framework to dynamic and multi-step manipulation scenarios.


\nocite{langley00}

\bibliography{AffordanceGrasp-R1}
\bibliographystyle{icml2026}

\newpage
\appendix
\onecolumn

\appendixcontentspage   

\section{CoT Generation Details}

\subsection{Prompt for CoT Generation}
\label{sec:cot_prompt}




\newtcolorbox{promptbox}[1]{  
        enhanced,
        toptitle=1.5mm,     
        bottomtitle=1.5mm,  
        arc=2mm,                  
        boxrule=0.6pt,            
        colback=white,            
        colframe=gray!90!black,   
        colbacktitle=gray!90!black, 
        coltitle=white,           
        fonttitle=\bfseries,      
        title=#1,                 
        left=5pt, right=5pt,      
        top=5pt, bottom=5pt,       
        fontupper=\footnotesize, 
}
\noindent
\begin{minipage}{\linewidth}

\begin{promptbox}{Prompt for Reasoning Sampling}  
    Analyze the following task step by step to derive the best possible answer.\\  
    Instruction: \{instruction\}\\  Ground Truth Answer (For verification): \{answer\}\\  
    IMPORTANT INSTRUCTIONS:\\
    Your goal is to explain HOW to solve this problem through visual analysis and reasoning. NOT to simply describe what the answer says.\\
    Please provide:\\
    \hspace*{0.5em}1. A detailed step-by-step reasoning process\\
    \hspace*{0.5em}2. Explain HOW you determine each part of the answer through visual observation \\
    \hspace*{0.5em}3. Verify the accuracy of your reasoning \\
    \hspace*{0.5em}4. Give your final answer clearly \\
    Focus on the REASONING PROCESS that explains HOW someone would solve this WITHOUT
    having access to the ground truth answer.
    \label{box:cot_sampling}
\end{promptbox}
\begin{promptbox}{Prompt for Summarizing Rules from Examples} 
    You are given affordance prediction reasoning examples. Your task is to extract GENERALIZABLE problem-solving principles that can be applied to NEW problems WITHOUT access to ground truth labels (bbox, point, aff\_methods, aff\_parts).\\
    Present as structured steps: \\
    \hspace*{0.5em}- Step 1: principle about scene and object analysis \\
    \hspace*{0.5em}- Step 2: principle about visual property recognition \\
    \hspace*{0.5em}- Step 3: principle about affordance inference \\
    \hspace*{0.5em}- Step 4: principle about part identification \\
    \hspace*{0.5em}- Step 5: principle about localization ... \\
    Critical Requirements: \\
    \hspace*{0.5em} Principles must work by analyzing the image and instruction only \\
    \hspace*{0.5em} Focus on how to solve through visual reasoning, not how to read answers \\
    \hspace*{0.5em} Don't write rules that reference "provided affordance name", "given part name", or "ground truth labels" \\
    Ensure principles are generalizable to similar but different problems
    \label{box:cot_summarizing1}
\end{promptbox}
\end{minipage}

    \begin{promptbox}{Prompt for CoT Generation with Extracted Rules}  
    Use the following principles to answer the question:\\
    \{rules\} \\
    Question: \{question\}\\ 
    Answer: \{answer\} \\
    Provide a concise solution with key reasoning steps in the following format: \\
    \texttt{<think>} "Reasoning Process"\texttt{</think>}\texttt{<answer>}"Final answer"\texttt{</answer>}
    \label{box:cot_summarizing2}
    \end{promptbox}

\begin{minipage}{\linewidth}
    \begin{promptbox}{Prompt for Verifying and Refining Reasoning and Answers}  

    \{response\} \\
    Please evaluate the structured response above for logical consistency and completeness. 
    Focus on the following aspects:\\
     \hspace*{0.5em}1. Does the reasoning in \texttt{<think>} logically support the conclusion in \texttt{<answer>}?\\
     \hspace*{0.5em}2. Does the reasoning include visual feature analysis and affordance analysis? \\
     \hspace*{0.5em}3. Are there any internal contradictions, logical errors, or missing key steps in the reasoning? \\ 
     \hspace*{0.5em}4. Is the reasoning chain complete and valid?\\
    Provide your evaluation in the following format:\\
     \hspace*{0.5em}\texttt{<reason>}A brief explanation of your assessment  \texttt{</reason>} 
     \hspace*{0.5em}\texttt{<validation>}Valid /Invalid\texttt{</validation>}
    Then, regardless of whether the reasoning is valid, output the full response again in the format below: \\
     \hspace*{0.5em}- Keep \texttt{<answer>} unchanged. \\
     \hspace*{0.5em}- Modify \texttt{<think>} only if necessary to ensure logical soundness and coherent reasoning. \\
     \texttt{<think>}final version of the reasoning process\texttt{</think>}
     \texttt{<answer>}original final answer\texttt{</answer>}
    \label{box:cot_verifying}
    \end{promptbox}
\end{minipage}

\section{Extended Related Work}

\subsection{Post-Training for MLLMs}
The post-training stage plays a crucial role in bridging the gap between pre-training and task adaptation for MLLMs~\cite{kumar2025llm}~\cite{chung2024scaling}. It typically consists of two major approaches: Supervised Fine-Tuning (SFT)~\cite{radford2021learning}~\cite{zhou2023lima}and Reinforcement Learning (RL)~\cite{zhai2024finetuning}. SFT generally relies on human-designed instruction response pairs or demonstration data to adapt the model, guiding it to follow specific formats or behaviors. Model families such as FLAN~\cite{wei2022finetuned} and LIMA~\cite{zhou2023lima} have demonstrated the effectiveness of SFT in task transferability and training stability. However, the performance of SFT is often constrained by the scale and quality of the training data, which limits its generalization ability.
In contrast, RL performs post-training by optimizing the model’s task strategies and behavioral quality through reward mechanisms. A typical approach is  Reinforcement Learning from Human Feedback(RLHF)~\cite{ouyang2022training}, which constructs reward signals based on human preferences or rule-based consistency, thereby improving the desirability and reliability of model outputs. Recent works such as the DeepSeek-R1 series~\cite{guo2025deepseek} adopt the Group Relative Policy Optimisation(GRPO) algorithm~\cite{shao2024deepseekmath} to strengthen reasoning capabilities: DeepSeek-R1-Zero removes SFT entirely and trains the model purely via RL, whereas DeepSeek-R1 introduces cold-start CoT-style examples to stabilize training, forming a more robust multi-stage reasoning optimization pipeline.

Importantly, for reasoning-oriented tasks, performing a cold-start SFT stage before RL greatly improves training stability during the subsequent RL phase~\cite{chu2025sft}~\cite{kumar2025llm}. SFT and RL are not competing approaches but complementary mechanisms. SFT provides the model with reasoning templates and task formats, while RL further refines output strategies through trial-and-error optimization. In summary, combining SFT and RL in a unified post-training pipeline allows the strengths of both approaches to be fully leveraged: SFT offers structural guidance and a stable initialization, whereas RL enhances the depth and quality of reasoning. Following this paradigm, our work applies an SFT + RL post-training process to improve MLLMs’ instruction understanding, environmental analysis, and task reasoning capabilities, enabling more reliable performance in complex affordance-centric scenarios.

\subsection{Chain-of-Thought in MLLMs}
Chain-of-Thought (CoT) reasoning has been widely demonstrated as a key strategy for enhancing the ability of large language models to handle complex tasks. By decomposing the reasoning process into a sequence of intermediate logical steps, CoT encourages models to generate interpretable step-by-step reasoning traces rather than only final answers. Numerous studies have shown that this mechanism not only significantly improves accuracy in tasks such as mathematical reasoning, logical inference, and multi-hop question answering, but also enhances model controllability and interpretability~\cite{wei2022chain}, as evidenced in models like PaLM~\cite{chowdhery2023palm} and the LLaMA-2~\cite{touvron2023llama}. Further work, such as Zero-Shot CoT~\cite{wang2023plan}, indicates that even simple prompts like ``\textit{Let’s think step by step}'' can induce coherent reasoning processes in large models. In addition, CoT-style data is often used for supervised fine-tuning, enabling models to learn continuous and structured reasoning procedures.

In the domain of vision-language models (VLMs), CoT reasoning has undergone a similar progression, expanding from text-based tasks to complex multimodal scene understanding. Recent studies demonstrate that constructing CoT data with visual explanatory steps – such as progressively describing scene structure, spatial relationships, or key object attributes – can significantly enhance a model's reasoning consistency in tasks like complex visual question answering and step-by-step scene interpretation. With the development of multimodal CoT approaches such as LLaVA-CoT~\cite{xu2025llava} and LlamaV-o1~\cite{thawakar2025llamav}, CoT has been successfully applied to the joint modeling of images, question answering, and reasoning. In affordance-related tasks, reasoning chains offer a natural advantage: the model typically needs to first perceive the scene, then identify relevant object parts, and finally infer affordance regions and potential actions. Therefore, inspired by the above CoT mechanisms, we construct a specialized reasoning dataset tailored for affordance tasks, enabling the model to learn object structure, functional relationships, and scene semantics through step-by-step reasoning. This approach ultimately enhances the model’s generalization ability and fine-grained reasoning performance in open-world affordance scenarios.

\section{Experimental Settings}
\subsection{Implementation Details}
\label{sec:implementation details}
Our method employs Qwen2.5VL-7B~\cite{bai2025qwen25vl} as the reasoning model and SAM 2-Large~\cite{ravi2024sam} as the segmentation model. Experiments are conducted on a server equipped with four NVIDIA A100 GPUs (80 GB each). All models are optimized using AdamW~\cite{loshchilov2019decoupled}. During the CoT cold-start stage, we use an initial learning rate of $1 \times 10^{-5}$ with a cosine annealing schedule and a batch size of 32 for one epoch. In the subsequent reinforcement learning stage, the learning rate is fixed at $1 \times 10^{-6}$ with a weight decay of 0.01; the batch size is set to 16, and 8 responses are sampled per rollout step.

\subsection{Evaluation Metrics}
In the following experiments, we report performance using the generalized Intersection-over-Union (gIoU) and complete Intersection-over-Union (cIoU) as evaluation metrics. These metrics are defined as follows:
\begin{subequations}
\begin{align}
    \mathrm{gIoU} &= \frac{1}{N_D} \sum_{i=1}^{N_D} \frac{|M_i \cap \hat{M_i}|}{|M_i \cup \hat{M_i}|}, \\
    \mathrm{cIoU} &= \frac{\sum_{i=1}^{N_D} |M_i \cap \hat{M}_i|}{\sum_{i=1}^{N_D} |M_i \cup \hat{M}_i|}.
\end{align}
\end{subequations}
where $N_D$ denotes the number of samples in the test set, $M_i$ represents the ground-truth mask of the $i$-th test sample, and $\hat{M}_i$ denotes  the corresponding predicted mask.

\begin{table*}[t!]
\centering
\caption{Test suites adopted from RAGNet~\cite{wu2025ragnet} for evaluating methods under in-domain, category-level zero-shot, and domain-level zero-shot settings, where HANDAL$^\dagger$ denotes a 1k subset of HANDAL.}
\begin{tabular}{lccccc}
\toprule
\textbf{Validation set} & \textbf{Instances} & \textbf{Zero-shot} & \textbf{Out-Domain dataset}& \textbf{Reasoning}  \\
\midrule
HANDAL~\cite{guo2023handal}  & 65351 & $\times$ &  $\times$  & $\times$ \\
HANDAL$^\dagger$~\cite{guo2023handal}       & 1003  & $\times$       &    $\times$    &  $\times$ \\
GraspNet-seen~\cite{fang2020graspnet}          & 1008  & $\times$       & $\times$   & $\times$   \\
GraspNet-novel~\cite{fang2020graspnet}         & 1018  & $\checkmark$ & $\times$  & $\times$ \\
3DOI~\cite{qian2023understanding}                   & 1012  & $\checkmark$   & $\checkmark$  & $\times$   \\
HANDAL$^\dagger$-easy~\cite{guo2023handal}  & 1003  & $\times$       & $\times$ &  $\checkmark$   \\
HANDAL$^\dagger$-hard~\cite{guo2023handal}  & 1003  & $\times$      & $\times$ & $\checkmark$  \\
3DOI~\cite{qian2023understanding}              & 1012  & $\checkmark$   & $\checkmark$ & $\checkmark$  \\
\bottomrule
\end{tabular}
\label{tab:testsets}
\end{table*}

To evaluate the performance of the proposed AffordanceGrasp-R1, we adopt the RAGNet benchmark~\cite{wu2025ragnet} and report results on eight test subsets spanning in-domain, category-level zero-shot, and domain-level zero-shot settings, as summarized in Tab.~\ref{tab:testsets}. We comprehensively evaluate AffordanceGrasp-R1 by comparing it against a diverse set of representative baselines, including VLPart~\cite{sun2023going} + SAM~2~\cite{ravi2024sam}, GroundingDINO~\cite{liu2024grounding} + SAM~2, Florence~2~\cite{xiao2024florence} + SAM~2, LISA-7B~\cite{lai2024lisa}, GLaMM~\cite{rasheed2024glamm}, and Qwen2.5VL-7B~\cite{bai2025qwen25vl} + SAM~2. We further include recent reinforcement learning–based approaches, namely Segzero~\cite{liu2025seg}, VisionReasoner~\cite{liu2025visionreasoner}, and Affordance-R1~\cite{wang2025affordance}. In addition, the current state-of-the-art (SOTA) method, AffordanceNet~\cite{wu2025ragnet}, is included for comparison.

\subsection{Real Robot Setup}
\label{sec:real_robot_Implementation}

To evaluate the open-world generalization capability of our approach in real-world manipulation scenarios, we conduct a series of language-conditioned robotic grasping experiments on a physical platform under a strict zero-shot protocol. This evaluation assesses the model’s ability to generalize to unseen objects and tasks without any environment-specific demonstrations. Our experimental setup consists of a KUKA LBR iiwa R820 robotic arm (Fig.~\ref{fig:grasp_pose_generation}) equipped with a Robotiq Adaptive 2F-85 two-finger gripper. An Intel RealSense D455 RGB-D camera is mounted on the end-effector to capture tabletop observations. We adopt a standard eye-in-hand RGB-D configuration to ensure reproducibility and stable depth geometry for 6-DoF grasp generation. Model inference and the control pipeline are executed on a workstation equipped with an NVIDIA RTX 4090 GPU (24 GB) and an AMD Threadripper Pro 7955WX CPU. Following the task design of AffordanceNet~\cite{wu2025ragnet}, we construct ten language-conditioned grasping tasks over a modified set of object categories, including a can, screwdriver, hammer, ball, scoop, pot, banana, glove, knife, and mouse. Notably, half of these tasks require accurate localization of fine-grained actionable regions—for example, grasping the handle of a hammer rather than its head—thereby stressing the model’s part-level semantic grounding capability under language constraints. Each task is executed ten times, and we report the average task success rate. We emphasize that all real-robot experiments strictly follow a zero-shot setting: no demonstration images or videos collected from this environment are used during training. For grasp pose generation, we follow the standard pipeline in GA-Grasp~\cite{cao2025geometryaware}. Specifically, we first generate a set of 6-DoF grasp candidates from the full-scene point cloud and then apply instruction-conditioned affordance masks to filter candidates, retaining only those whose contact regions lie within the predicted actionable area. Finally, we compare our proposed AffordanceGrasp-R1 with AffordanceNet under identical hardware and evaluation protocols to validate the effectiveness of our vision-language-driven grasping framework.
\section{3D IoU Matching }\label{sec:3D_IoU_Matching}
\begin{figure}[H]
  \centering
  \includegraphics[width=0.5\linewidth]{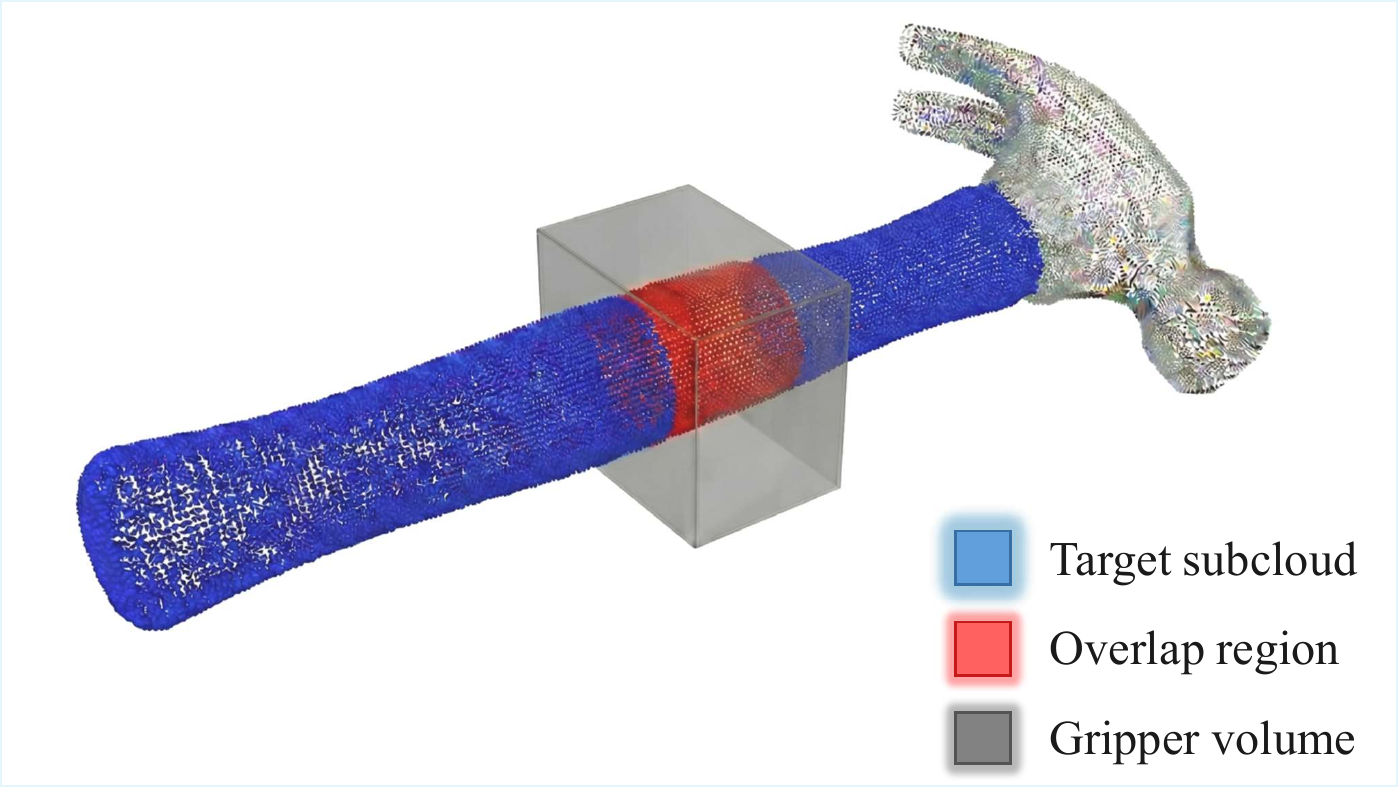}
   \caption{3D IOU matching for grasp selection: We compute the 3D IoU between the gripper closing volume and the semantics-aligned target region induced by the predicted affordance subcloud $\hat{S}$. The red region indicates the overlap volume used to compute the 3D IoU score.}
  \label{fig:3d_iou_matching}
\end{figure}

\section{Qualitative Results}\label{sec:experiment_results}

\begin{figure}[H]
    \centering
    \includegraphics[width=0.9\linewidth]{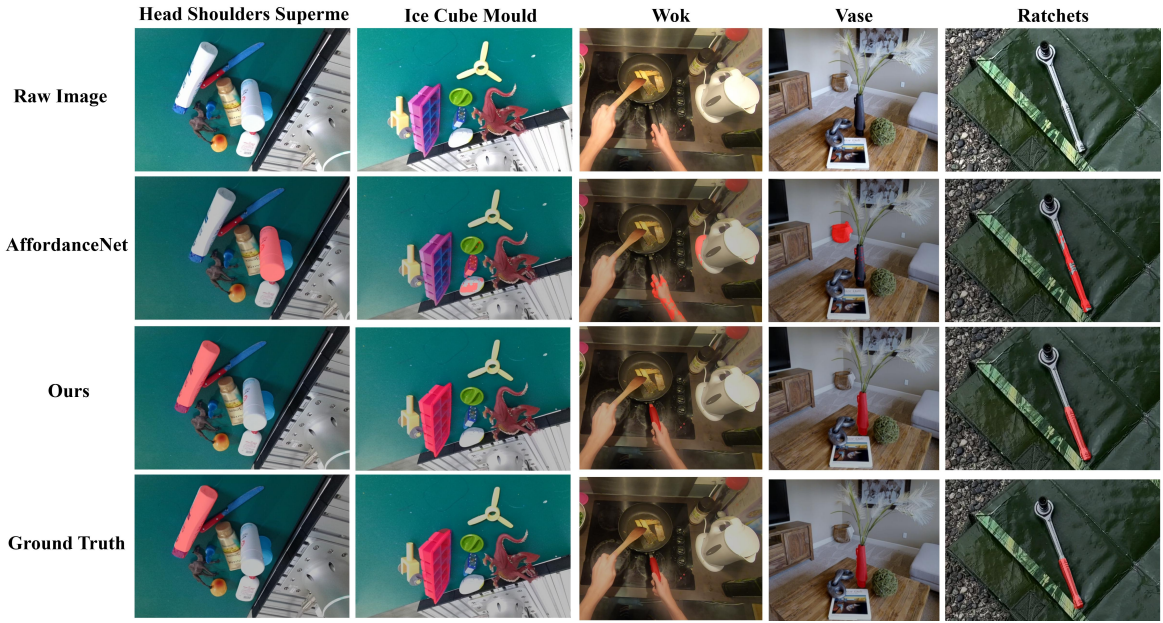} 
    \caption{Qualitative comparison of affordance segmentation results on the main subset~\cite{wu2025ragnet}.}
    \label{fig:affordance_visual_results}
\end{figure}

\begin{figure}[H]
    \centering
    \includegraphics[width=\linewidth]{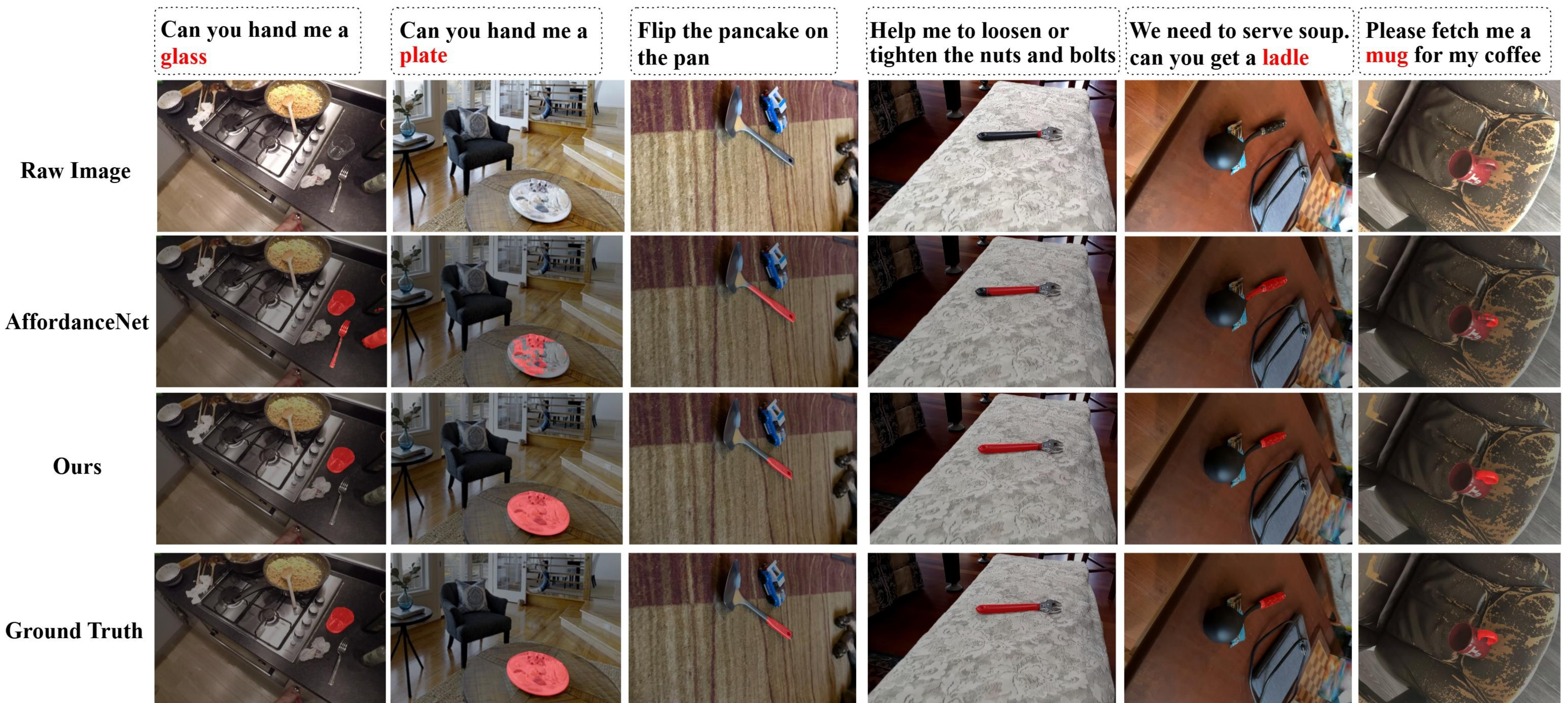} 
    \caption{Qualitative comparison of affordance segmentation on the reasoning-based subset. AffordanceGrasp-R1 consistently outperforms AffordanceNet in segmentation accuracy.}
    \label{fig:reasoning__visual_results}
\end{figure}

\begin{figure}[H]
    \centering
    \includegraphics[width=\linewidth]{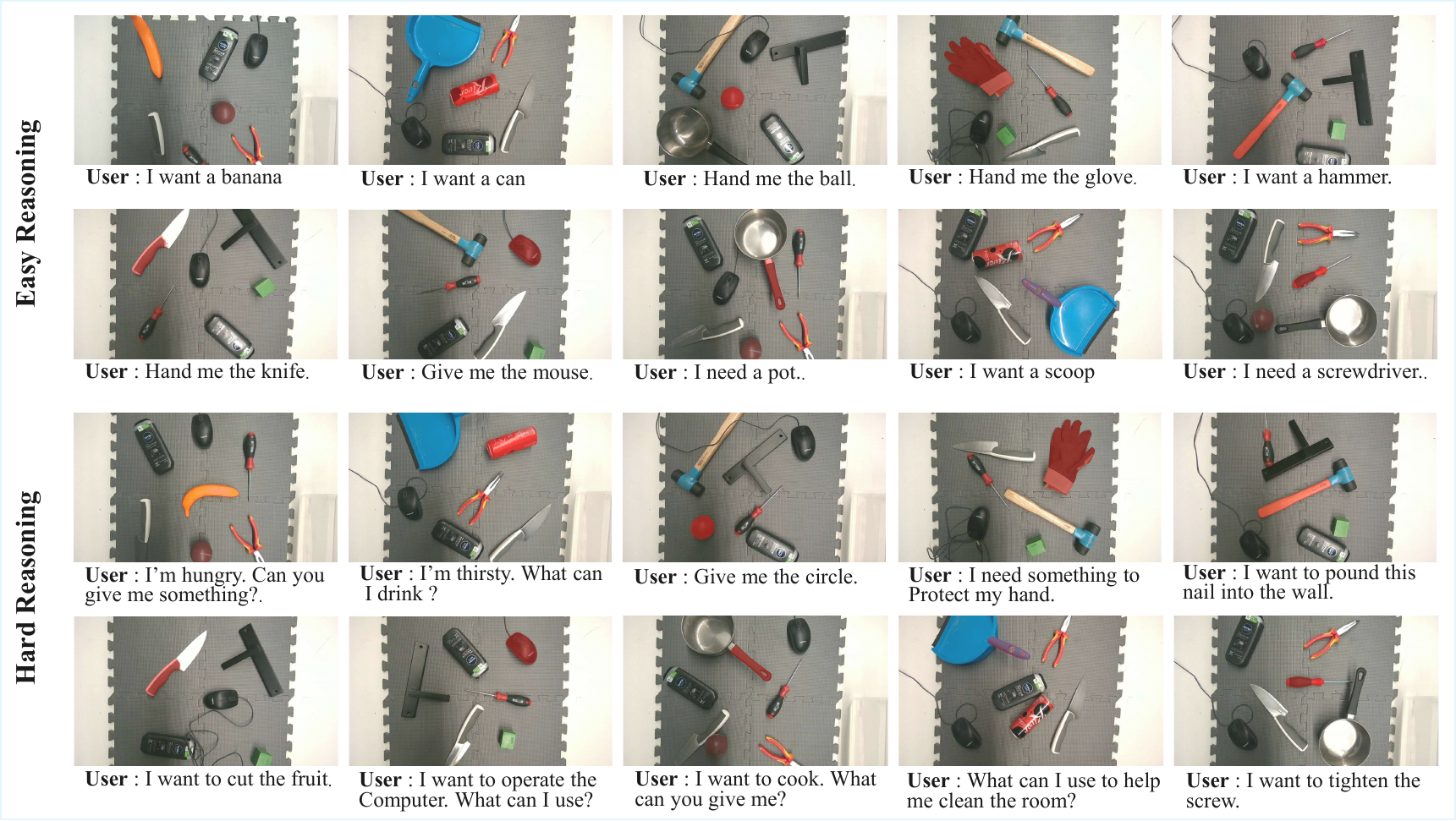} 
    \caption{More experiment results from our model. We use easy reasoning-based and hard reasoning-based instructions, respectively.}
    \label{fig:easy_hard_examples}
\end{figure}

\newcommand{\grouptitle}[1]{%
  \vspace{0.6mm}
  \noindent\textbf{\small #1}\par
  \vspace{0.4mm}
}

\begin{figure}[H]
    \centering
    \grouptitle{(A) Affordance-part grounding required }
    \includegraphics[width=\linewidth]{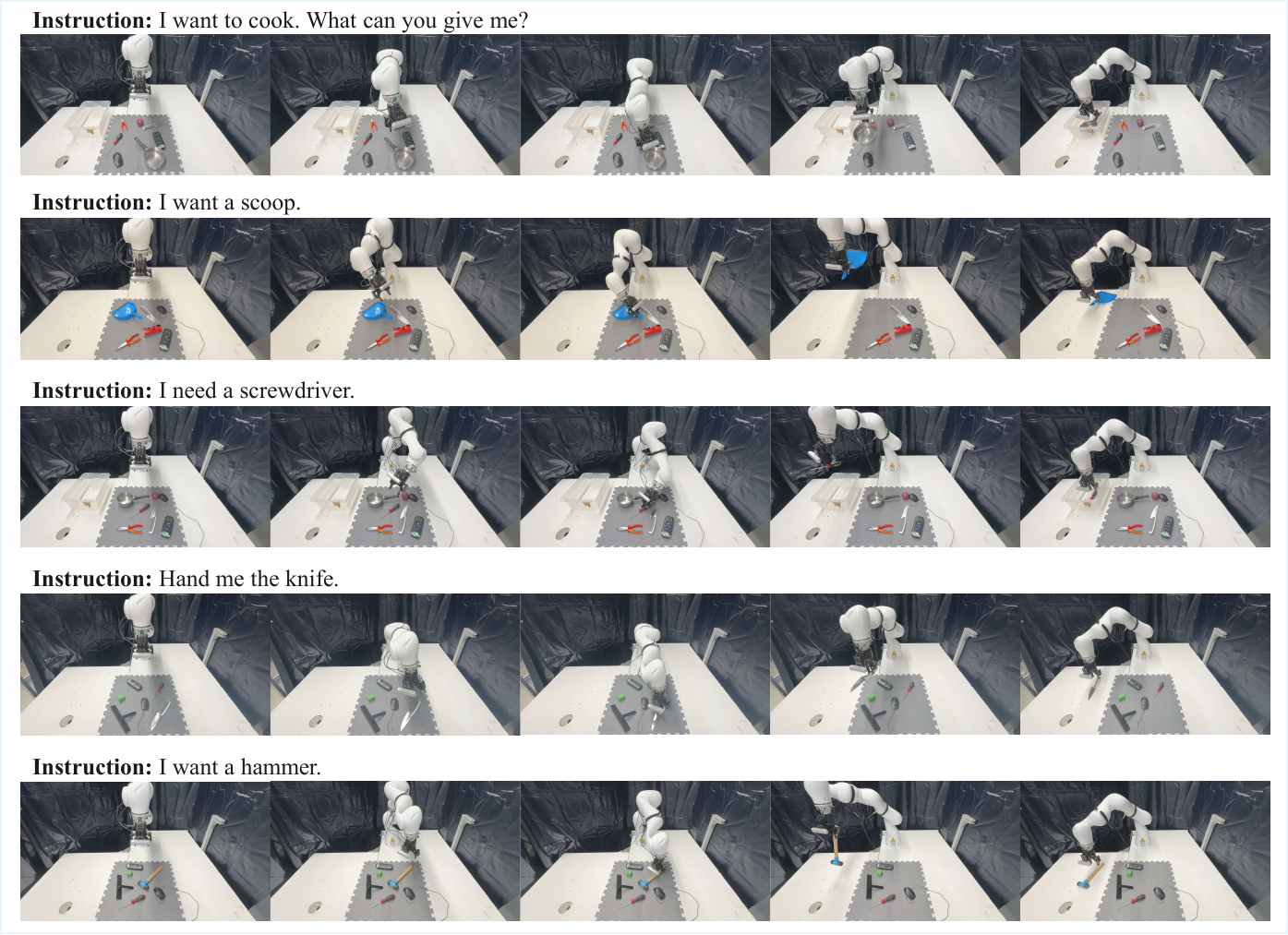} 
    \caption{Real-world grasping rollouts of AffordanceGrasp-R1 on a KUKA LBR iiwa with a Robotiq 2F-85 gripper. (A) Tasks that require object-centric affordance-region reasoning.}
    \label{fig:rollout_affordance}
\end{figure}

\begin{figure}[H]
    \centering
    \grouptitle{(B) Object-level grounding (no part constraint) }
    \includegraphics[width=\linewidth]{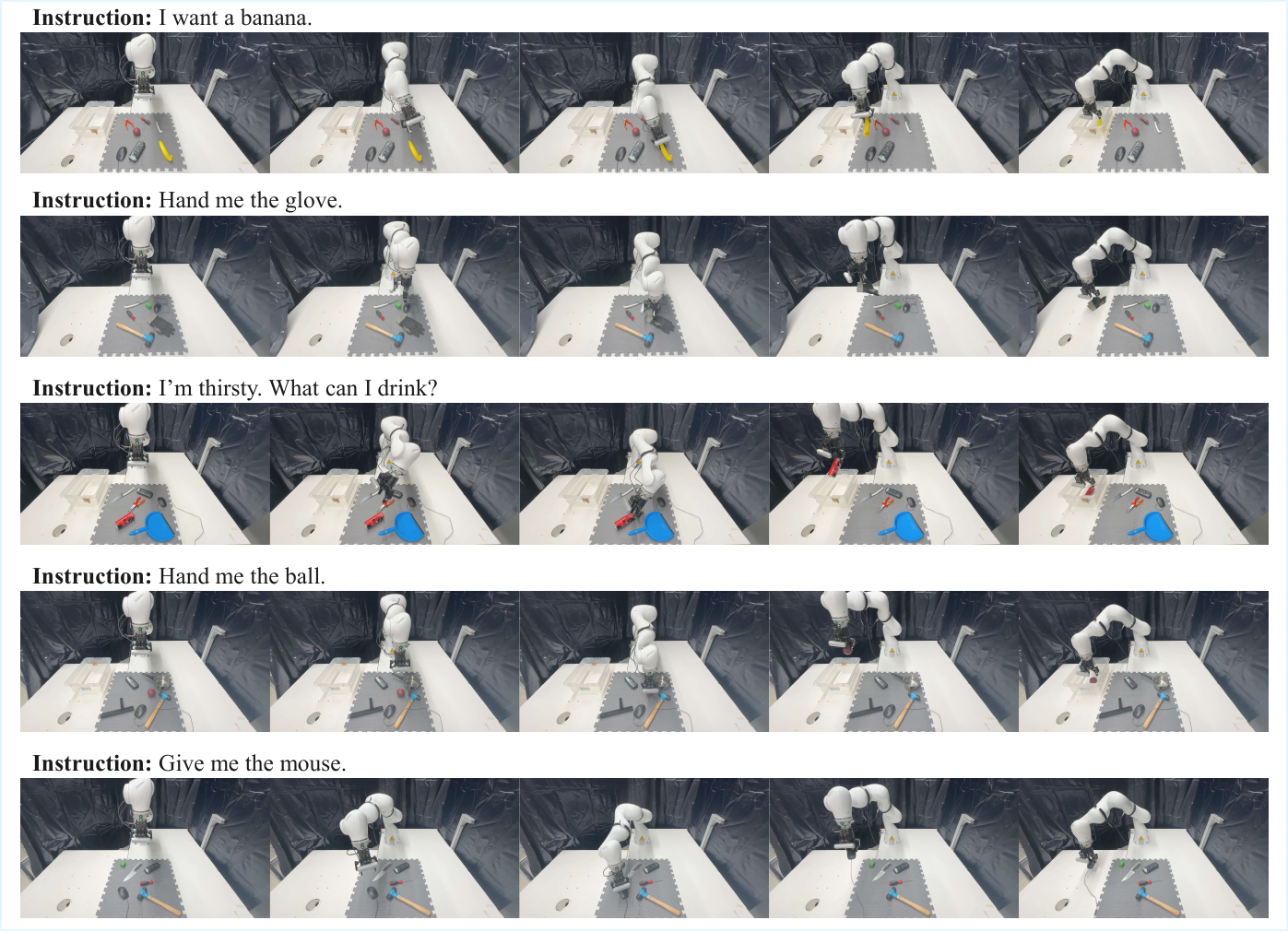} 
    \caption{Real-world grasping rollouts of AffordanceGrasp-R1 on a KUKA LBR iiwa with a Robotiq 2F-85 gripper. (B) Tasks where object-level grounding is sufficient and no object-part affordance reasoning is needed for successful grasping.}
    \label{fig:rollout_direct}
\end{figure}

\end{document}